\documentclass{bmvc2k}
\usepackage{amssymb,amsmath,dsfont}
\usepackage{multicol}
\usepackage{multirow}
\usepackage{float}
\usepackage{ifthen}
\usepackage{bbm}
\usepackage{bm}
\usepackage{algorithm}
\usepackage{hyperref}

\newcommand{\method}[1]{\ifthenelse{\equal{#1}{full}}{Local-Global Associative Assembling}{LOGA}}
\newcommand{\best}[1]{{\color{red}\textbf{#1}}}
\newcommand{\scnd}[1]{{\color{blue}\underline{#1}}}

\renewcommand{\paragraph}[1]{\vspace{0.1cm}\noindent\textbf{#1}\quad}

\def\ie{\emph{i.e}\bmvaOneDot}
\def\vs{\emph{v.s}\bmvaOneDot}
\def\eg{\emph{e.g}\bmvaOneDot}

\graphicspath{{./figures/}}

\title{Local-Global Associative Frame Assemble in Video Re-ID}

\addauthor{Qilei Li}{q.li@qmul.ac.uk}{1}
\addauthor{Jiabo Huang}{jiabo.huang@qmul.ac.uk}{1}
\addauthor{Shaogang Gong}{s.gong@qmul.ac.uk}{1}

\addinstitution{
 Computer Vision Group, \\
 School of Electronic Engineering and Computer Science, \\
 Queen Mary University of London, \\ London E1 4NS, UK
}

\runninghead{Q. Li, J. Huang, S. Gong}{Local-Global Associative Frame Assemble}

\begin{document}
\maketitle

\begin{abstract}
Noisy and unrepresentative frames in 
automatically generated object bounding boxes from video sequences 
cause significant challenges 
in learning discriminative representations 
in video re-identification (Re-ID). 
Most existing methods tackle this problem by 
assessing the importance of video frames 
according to either their local part alignments 
or global appearance correlations separately. 
However, given the diverse and unknown sources of noise which usually co-exist in captured video data, 
existing methods have not been effective satisfactorily. 
In this work, we explore jointly both local alignments and global correlations
with further consideration of their mutual promotion/reinforcement
so to better assemble complementary discriminative Re-ID information 
within all the relevant frames in video tracklets. 
Specifically, we concurrently optimise 
a local aligned quality (LAQ) module that distinguishes the quality of each frame based on local alignments, 
and a global correlated quality (GCQ) module that estimates global appearance correlations. 
With the help of a local-assembled global appearance prototype,
we associate LAQ and GCQ to exploit their mutual complement.
Extensive experiments demonstrate the superiority of the proposed model 
against state-of-the-art methods 
on five Re-ID benchmarks, 
including MARS, Duke-Video, Duke-SI, iLIDS-VID, and PRID2011.

\end{abstract}

\vspace{-0.5em}
\section{Introduction}
\label{sec:intro}

\noindent
Person re-identification (Re-ID) 
aims to match pedestrian's identity
across disjoint cameras views distributed at different
locations~\cite{chen2018video,wu2020decentralised,wang2014person,wu2021generalising}. 
Early Re-ID studies concentrated on
exploring appearance patterns unique per identity from still images~\cite{fu2019horizontal,li2018harmonious,yao2019deep},
which has shown remarkable discrimination capacity.
However,
such methods assume well-curated data and the identity information are preserved in images.
This assumption dramatically restricts their scalability and usability 
to many practical application scenarios when uncontrollable
environments are the norm not the exception where video data are captured~\cite{li2019unsupervised, liu2017quality}.
Video person Re-ID beyond still images requires analysing and assembling
information from a sequence of video frames in each tracklet so to
build a more discriminative and robust representation of pedestrians
in motion, minimising information corruption from poor frames and
ID-switch~\cite{liu2019spatial,xu2017jointly,chen2018video,fu2019sta,zhang2018multi, hou2020temporal}.

\begin{figure}[tbp]
	\centering
	\includegraphics[width=\textwidth]{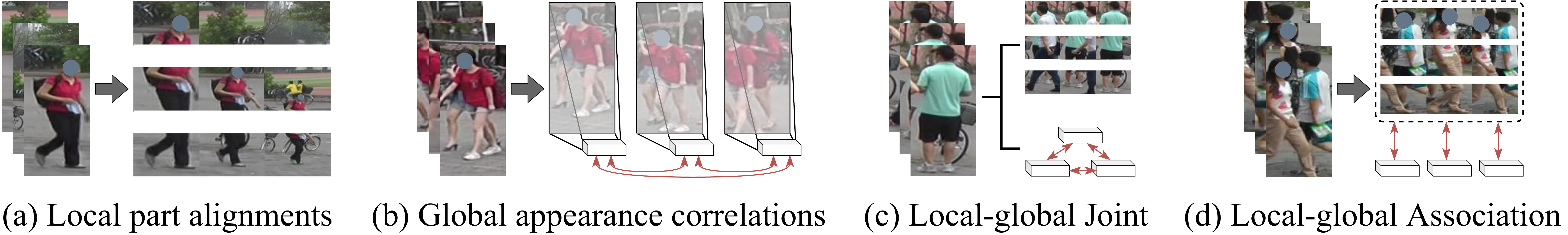}
 	\vspace{-1em}
	\caption{Illustration of four types of quality assessment strategies
	for frame assembling.}
 	\vspace{-1.5em}
	\label{fig:method_compare}
\end{figure}

In the literature,
one of the most commonly adopted techniques for
assembling identity information from different video frames
is {\em averaging} by pooling~\cite{suh2018part, si2018dual}.
By assuming all the frames are in equal importance,
the pooling method neglects their diverse qualities 
caused by the constantly changing environments 
and/or unreliable pedestrian detections. Therefore, the aggregated tracket's representations 
are likely impacted by various types of noise
as shown in Fig.~\ref{fig:method_compare}.
In order to {\em selectively} assemble video frames rather than averaging,
attention mechanisms~\cite{hu2018squeeze,vaswani2017attention,wang2018non,woo2018cbam,huang2021cross}
have been studied to explore the correlations between 
the \textit{global} visual features of frames (Fig.~\ref{fig:method_compare}~(b))
so that the common appearance patterns shared among frames in the same
tracklet are maintained while removing/ignoring unusual and
low-quality frames~\cite{li2020relation,matiyali2020video,liu2021watching,wang2021robust}. 
In contrast to the global appearance correlations,
an alternative approach~\cite{zhang2020multi,hou2019vrstc,hou2020temporal}
compares video frames by \textit{local} parts (Fig.~\ref{fig:method_compare}~(a))
so to identify outliers that are significantly misaligned with other
frames in a tracklet. Although sharing the same objective
to adaptively assemble only the relevant video frames, these two approaches 
differ in exploiting information in different granularities. In
isolation, both are sub-optimal in different real-world video scenes.
The local-parts approach is fragile if the detected pedestrians are not well-aligned
while the global-appearance approach is spatially insensitive, tending
to miscorrelate patterns of interest in the background. Beyond attentive assembling,
Recurrent Neural Network (RNN)~\cite{xu2017jointly,mclaughlin2016recurrent}
has also been exploited for modelling temporal information to
represent frame sequences in video tracklets. However, this approach
is also vulnerable to noisy frames without careful frame selections~\cite{wu2019spatio}.

In this work, we propose a tracklet frame assembling approach
to video person Re-ID termed \textit{\method{full}} (\method{abbr}).
As shown in Fig.~\ref{fig:method_compare}~(d),
the \method{abbr} method
adaptively assembles video frames
in the same tracklets
by a Local Aligned Quality (LAQ)
and a Global Correlated Quality (GCQ) modules
to assess importance/relevance of the frames by both their alignments in local part
and global appearance correlations
as well as their mutual reinforcements.
Whilst the focus of most existing spatial-temporal attentive methods
is on collaborating
the temporal information with \textit{intra-frame} spatial attention,
we aim to exploit the \textit{inter-frame} complements more effectively,
which is different 
and ready to benefit from the advancing per-frame learning.
Specifically,
the LAQ module
divides all video frames in a tracklet
into a same set of spatial parts and
assesses each frame's quality by their part-wise alignment to the
other frames so to measure both inter-frame visual similarity and spatial alignment.
On the other hands,
the GCQ module 
is applied on the holistic feature representation of each frame
to consider inter-frame global appearance correlations,
which is more robust to local part misalignment but spatially
insensitive so less reliable from miscorrelation of information, 
\eg irrelevant patterns in the background. 
Furthermore,
to associate the local and global information
and exploit their mutual benefits,
we take the tracklet's representation assembled by the LAQ
as its prototype
and compare the global visual feature of frames with it in the GCQ module
so that the two modules are encouraged to find a trade-off 
between the local and global information
to cope with different types of noise more reliably.

Contributions of this work are three-fold: (1) 
To our best knowledge,
we make the first attempt 
to explore the \textit{association and mutual promotion} 
of frame's local part alignments
and global appearance correlations
in assembling a sequence descriptor 
so to improve the model's robustness to noisy frames and inter-frame
ID-switch in video Re-ID.
(2) We propose a new video person Re-ID model termed 
\textit{\method{full}} (\method{abbr})
that learns a discriminative and reliable representation for video tracklets
by adaptively assembling frames of diverse qualities.
(3) We introduce a local-assembled global appearance prototype
to \textit{associate} the local and global visual information
by exploiting their mutual agreements to facilitate the learning of a discriminative tracklet representation.

Extensive experiments show the performance advantages and
superior robustness of the proposed \method{abbr} model over the
state-of-the-art video Re-ID models on four video Re-ID
benchmarks MARS~\cite{zheng2016mars},
Duke-Video~\cite{ristani2016performance, wu2018exploit},
Duke-SI~\cite{li2019unsupervised}, and iLIDS-VID~\cite{wang2014person}. 

\vspace{-0.5em}
\section{Related Works}
\label{sec:related}

\noindent
Video person Re-ID aims to learn an expressive appearance feature 
and/or distance metric from a sequence of frames, 
\ie, a video tracklet.
To take the advantages of the additional temporal information
and complementary spatial information intrinsically available in video tracklets,
existing approaches explore either local part alignments~\cite{song2018region,bao2019preserving,zhang2020multi,hou2019vrstc,hou2020temporal}
or global appearance correlations~\cite{liu2017quality,li2019global,zhang2019scan,li2020temporal,li2020relation,matiyali2020video,liu2021watching,wang2021robust}
to assemble the per-frame representations
with high robustness to their diverse qualities.

\paragraph{Local part alignments.}
Considering the consistent body structure shared among humans
and the arbitrary combinations of body part's appearance 
that unique to each identity,
it is intuitive to differentiate images/frames of pedestrians
regarding their visual similarity in different parts.
In this spirit,
local-parts assembling approaches~\cite{song2018region,bao2019preserving,zhang2020multi,hou2019vrstc,hou2020temporal} apply per-part comparisons of video frames in the same tracklets
to identify outliers which are misaligned with others in most local parts,
so as to
restore the corrupted parts of frames with the complements of others~\cite{hou2019vrstc,hou2020temporal}
or
degrade their importance in frame assembling~\cite{song2018region, bao2019preserving,zhang2020multi}.
However,
this hypothesis that a pedestrian detected in different video frames
being mostly well-aligned is often untrue due to 
unreliable auto-generated person bounding boxes, %
\eg the importance of a noise-free video frame 
might be underestimated
due to the spatial shift of its detected bounding box 
from those in other frames.
In this work,
we further consider the holistic visual similarity of video frames
when assessing their quality,
which helps refrain from inaccurate assessments caused by part misalignments. 

\paragraph{Global appearance correlations.}
In contrast to the local-parts approaches,
methods based on global-appearance~\cite{liu2017quality,li2019global,zhang2019scan,li2020temporal,li2020relation,matiyali2020video,liu2021watching,wang2021robust}
take the advantages of 
the strong representational power of convolutional neural network (CNN)~\cite{goodfellow2016deep,lecun2015deep}
to learn correlations between video frames holistically
so that the irrelevant frames, which are likely in low-quality, 
are suppressed in frame assembling.
However, the CNN features can be insensitive to spatial shift
resulting in potential miscorrelations of visually similar but
irrelevant parts, 
\eg the ID-switch issue shown in Fig.~\ref{fig:method_compare} (b) 
is hard to be detected
due to the subtle differences in the two pedestrians' outfits.
This will result in misassemblling of frames to represent a
tracklet. To address this problem, we propose to enhance the global-appearance methods
by jointly explore frames' holistic visual correlations
and their local part alignments by considering inter-frame spatial relations.

\paragraph{Spatial attention.}
Beyond the temporal assembling approaches discussed above,
spatial attention~\cite{woo2018cbam} is also popular 
in both image and video person Re-ID~\cite{li2018harmonious,zhong2020robust,xiang2020part,fu2019sta,wu2019spatio}.
By exploring the correlations of local parts within a still image
or across different video frames,
the spatial attention mechanism
is able to adaptively focus on the more discriminative regions
regardless of their spatial location.
However, 
this is prone to miscorrelation of information 
in video frames
as in the global-correlated assembling approaches.
Differently, our LAQ module investigates the alignments of 
the same part across different video frames,
focusing on exploiting complementary inter-frame information in a tracklet.

There are a few recent attempts on exploring
jointly the local and global information for frames assembling in video Re-ID~\cite{chen2020frame,yang2020spatial}.
However,
they learn from these two types of information with few interactions
either by a dual-branch network~\cite{chen2020frame} or feature concatenations~\cite{yang2020spatial},
and overlook the local-global mutual impacts
(Fig.~\ref{fig:method_compare}~(c)). 
We validated the effectiveness of the proposed \method{abbr} over 
those assembling strategies
in both performance evaluation (Section~\ref{sec:SOTA})
and ablation analysis (Section~\ref{sec:ablation}).

\vspace{-0.5em}
\section{Video Person Re-ID}

Given $N$ video tracklets $\mathcal{T} = \{\bm{T}_i\}_{i=1}^N$ with each containing
$L$ frames $\bm{T}_i = \{\bm{I}^i_j\}_{j=1}^L$ depicting 
$C$ pedestrians in motion, 
the objective of video person Re-ID is to
derive a representation model $\theta$ from the tracklets data $\mathcal{V}$
which is capable of extracting discriminative feature representations $\bm{x}$: 
$f_\theta(\bm{T}) \rightarrow \bm{x}$
for Re-ID matching across disjoint camera views.
Considering the diverse and unknown sources of noise
commonly exist in surveillance videos,
which leads to distractions in different frames,
it is essential for the model
to effectively recognise visual patterns that specific to each pedestrian
to selectively assemble frames into a tracklet's representation.
This is inherently challenging
due to the uncertain nature of noise in tracklets of
people in motion against backgrounds of visually similar distractors.

\subsection{Local-Global Associative Assembling}

\begin{figure}
    \centering
    \includegraphics[width=0.9\textwidth]{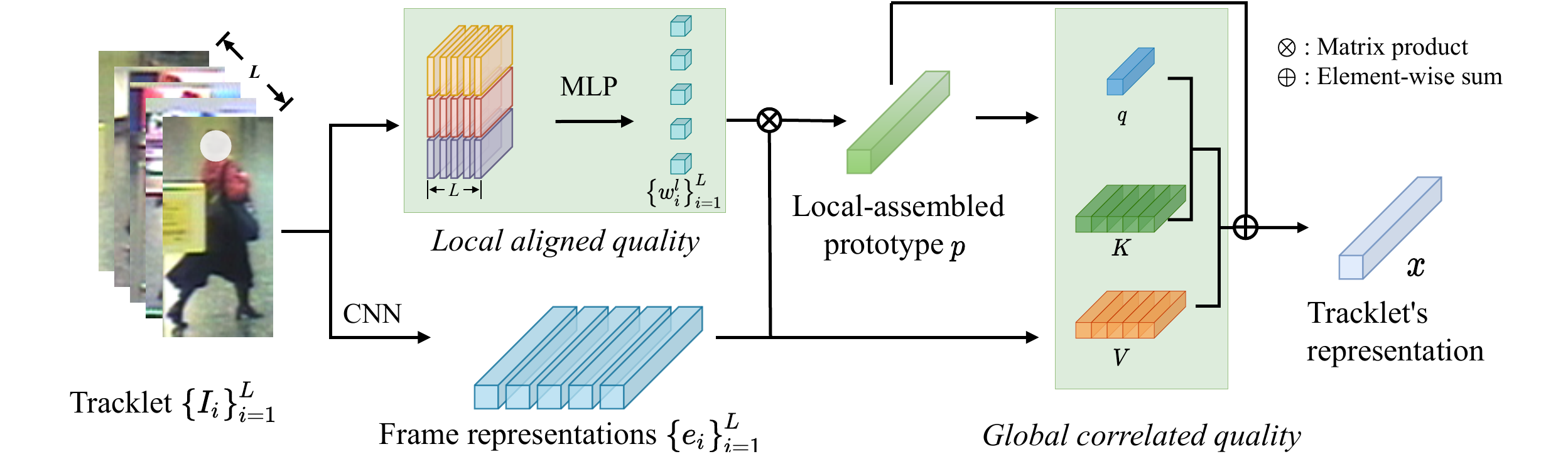}
    \caption{Overview of the proposed 
    \textit{\protect\method{full}} (\protect\method{abbr}) method.
   }
	\vspace{-1em}
    \label{fig:framework}
\end{figure}

In this work,
we propose a \textit{\method{full}} (\method{abbr}) model
to address this problem by
selecting information from video frames in the same tracklets
according to both their 
local part alignments and global appearance correlations
as well as the synergy and mutual promotion of these two types of information.
For notation clarity,
in the following, we focus on the formulation of assembling frames $\{\bm{I}_i\}_{i=1}^L$ 
in a single video tracklet $\bm{T}$
and ignore its tracklet index.
As shown in Fig.~\ref{fig:framework},
the video tracklet is first fed into
a \textit{Local Aligned Quality} (LAQ) module
to assess the quality of frames
regarding their part-wise alignment:
\begin{equation}
\{w^l_i\}_{i=1}^L=f_{\theta_l}(\{\bm{I}_i\}_{i=1}^L).
\label{eq:laq}
\end{equation}
The $\theta_l$ in Eq.~\eqref{eq:laq} is the learnable parameters
of the LAQ and $w^l_i$ denotes the importance of frames $\bm{I}_i$
determined by its alignments with other frames in local parts.
Then,
a \textit{global correlated quality} (GCQ) module is devised
which is applied to the $D$-dim
holistic visual representation
$\bm{E} = \{\bm{e}_i\}_{i=1}^L \in \mathbb{R}^{D\times L}$
of frames to determine their global appearance correlations.
Instead of focusing on only the global visual features
that are prone to spatial-insensitive miscorrelation,
we explore the mutual synergy between local and global information
by associating LAQ and GCQ through a prototypical descriptor $\bm{p}$.
This assembles a frame's global features by their local-parts quality
in GCQ for correlation exploration:
\begin{align}
\bm{p} &= \sum_{i=1}^L w^l_i\bm{e}_i, \label{eq:prototype} \\
\{w^g_i\}_{i=1}^L &= f_{\theta_g}(\{\bm{e}_i\}^L_{i=1} \vert \bm{p}),
\label{eq:gaq}
\end{align}
where $\{w_i^g\}_{i=1}^L$ denotes 
frame's quality 
regarding their global-appearance feature $\bm{E}$
and $\theta_g$ is the learnable parameters of the GCQ module.
In this way,
the final representaion $\bm{x}$ of a tracklet $\bm{T}$
is obtained by associating LAQ and GCQ through $\bm{p}$:
\begin{equation}
\bm{x} = f(\bm{E} \vert \bm{w}^l;\bm{w}^g).
\label{eq:assemble}
\end{equation}

With the tracklet-level representations,
the \method{abbr} model can be trained
with arbitarily conventional Re-ID objectives
in an end-to-end manner.
In inference,
a generic distance metric (\eg cosine distance)
is used to measure pairwise visual similarity of tracklets
for video Re-ID matching.
The overall learning process 
of the \method{abbr} model
is depicted in Algorithm~\ref{alg}.

\begin{algorithm}[ht]
    \caption{Local-Global Associative Assembling (LOGA).} \label{alg}
    \textbf{Input:} 
    Video tracklets $\mathcal{T}$,
    Identity labels $\mathcal{Y}$. \\
    \textbf{Output:}
    A deep CNN model for video person Re-ID. \\
    \textbf{for} $i=1$ \textbf{to} $max\_iter$ \textbf{do}
    \\\hphantom{~~}
    Randomly sample a mini-batch of video tracklets from $\mathcal{T}$ and their identity labels from $\mathcal{Y}$.
    \\\hphantom{~~}
    Compute the local-aligned per-frame importance scores (Eq.~\eqref{eq:laq}).
    \\\hphantom{~~}
    Feed the tracklets into backbone network to obtain their holistic visual features $\bm{E}$.
    \\\hphantom{~~}
    Compute the local-assembled global appearance prototype (Eq.~\eqref{eq:prototype}).
    \\\hphantom{~~}
    Compute the global-correlated per-frame importance scores (Eq.~\eqref{eq:gaq}).
    \\\hphantom{~~}
    Compute the tracklet-level representations (Eq.~\eqref{eq:assemble}).
    \\\hphantom{~~}
    Compute the objective losses and update the network by back-propagation.
    \\\textbf{end for}
\end{algorithm}

\paragraph{Local aligned quality.}
To explore the visual similarity of frames in terms of their local alignments,
we separate them uniformly into $M$ non-overlapping patches (parts)
and apply patch-wise cross-frame convolution
to recognise the aligned local patterns.
This is accomplished by
first flatten the 2D frames $\{\bm{I}_i\}_{i=1}^L$ 
then stacking them in the channel dimension
as the raw representation of the tracklet $\bm{T}$
maintaining the inter-frames spatial correspondence.
An 1D convolution is then applied on $\bm{T}$
to explore the per-part visual patterns,
\begin{equation}
\tilde{\bm{w}}^l = \bm{F} \ast \bm{T},\quad\bm{F} \in \mathbb{R}^{S \times L \times L},
\label{eq:laq_conv}
\end{equation}
where $\ast$ denotes the 1D convolution function and
$\bm{F}$ is a trainable kernel.
The size $S$ of kernel $\bm{F}$ is determined by 
the granularity of the spatial separation,
\ie, $S=\frac{H\times W}{M}$ 
where $H$ and $W$ are the height and width of frames, respectively.
The computed results $\tilde{\bm{w}}^l \in \mathbb{R}^{M\times L}$ 
encode the part-wise importance of every frame,
which is then aggregated by pooling followed by a multi-layer perceptron (MLP)
to obtain the per-frame scores:
\begin{equation}
\bm{w}^l = f_{\theta_l}(\{\bm{I}_i\}_{i=1}^l) = \text{Softmax}(\text{MLP}(\text{Pooling}(\tilde{\bm{w}}^l))) \in (0,1)^{L\times 1}. \\
\label{eq:laq_mlp}
\end{equation}
The $\text{Pooling}(\cdot)$ in Eq.~\eqref{eq:laq_mlp} 
is a frame-wise mean pooling function
and the $\text{MLP}(\cdot)$
stands for
a single layer MLP activated by a ReLU function.
The resulted scores
are then normalised by softmax function %
as the indication $\bm{w}^l$ of per-frame importance to the tracklet $\bm{T}$.
In this way,
the LAQ learns to assess the frame's quality by its local part
alignments to other frames, so to identify the misaligned outlier frames
and suppress them from representing a tracklet.

\paragraph{Global correlated quality.}
The GCQ module is formulated to explore the inter-frame correlations
according to their global appearances.
However,
the spatial invariant characteristic of the CNN features
tends to miscorrelate patterns of interests with potential noise in the background,
\ie completely ignoring the spatial part's alignment.
In this case,
we propose to establish the GCQ on the results yielded by LAQ
so to associate them by their synergy.
Specifically,
given the frame's importance $\bm{w}^l$ computed by Eq.~\eqref{eq:laq_mlp}
regarding their local part alignments,
we first assemble their visual features accordingly
in Eq.~\eqref{eq:prototype},
which serves as the appearance prototype $\bm{p}$
of a tracklet. 
Then, the global-appearance quality of a frame is estimated according to
the correlation between their global features and the prototype:
\begin{equation}
\begin{gathered}
\bm{q} = f_{\theta_q}(\bm{p}) \in \mathbb{R}^{D\times 1},
\quad \bm{K} = f_{\theta_k}(\bm{E}) \in \mathbb{R}^{D\times L} \\
\bm{w}^g = f_{\theta_g}(\{\bm{e}_i\}_{i=1}^L \vert \bm{p}) = \text{Softmax}(\bm{K}^\top\bm{q}) \in (0, 1)^{L\times 1}.
\end{gathered}
\label{eq:gaq_scores}
\end{equation}
The $f_{\theta_q}$ and $f_{\theta_k}$ functions in Eq.~\eqref{eq:gaq_scores}
are to linearly transform respectively the prototype and frame's
features. Both are followed by batch normalisation.
In this way,
the video frames in $\bm{T}$
with higher appearance correlations
to the pedestrian's prototype $\bm{p}$
will be highlighted with larger $w^g_i$
and those mis-correlated ones will be suppressed.

\paragraph{Tracklet-level representation.}
Given the global-appearance quality of frames,
their visual features can be selectively aggregated by:
\begin{equation}
\bm{V} = f_{\theta_v}(\bm{E}) \in \mathbb{R}^{D\times L},\quad
\bm{\hat{p}} = \bm{V}{\bm{w}^g} \in \mathbb{R}^{D\times 1},
\label{eq:gaq_assemble}
\end{equation}
where $f_{\theta_v}$ is identical to 
$f_{\theta_q}$ and $f_{\theta_k}$ in Eq.~\eqref{eq:gaq_scores}
with independent parameters $\theta_v$.
Rather than taking $\bm{\hat{p}}$ 
as the final representation of the tracklet $\bm{T}$,
in light of the residual learning~\cite{he2016deep},
we distill the complementary information 
from global appearance correlations of frames to enhance 
the prototype computed by local-parts quality 
so to minimise representational error 
from identity-irrelevant part misalignments.
To that end,
we further learn the residual of $\bm{p}$ from $\bm{\hat{p}}$ 
and obtain the visual feature representation of $\bm{T}$ by:
\begin{equation}
\bm{x} = f(\bm{E} \vert \bm{w}^l;\bm{w}^g) = \bm{p} + \text{FC}(\bm{\hat{p}}) \in \mathbb{R}^{D\times 1}.
\label{eq:laq_gaq}
\end{equation}
This design 
not only explores the global features of frames
but also considers their local part alignments for 
optimising a discriminative tracklet representation.

\subsection{Model Training}

Given the formulations of LAQ and GCQ,
the proposed \method{abbr} model can benefit from conventional learning supervisions.
Specifically,
the \method{abbr} model is jointly trained with 
a softmax cross-entropy loss $\mathcal{L}_\text{id}$ 
and a triplet ranking loss $\mathcal{L}_{\text {trip}}$~\cite{hermans2017defense}. 
The softmax cross-entropy loss $\mathcal{L}_\text{id}$ is employed to 
optimise identity classification:
\begin{equation}
\tilde{\bm{y}}_i = \text{Softmax}(\text{FC}(\bm{x}_i)),\quad
\mathcal{L}_\text{id}(\bm{T}_i) = - \sum_{j=1}^C y_{i,j} \log \tilde{y}_{i,j}.
\label{eq:loss_id}
\end{equation}
The $\bm{y}_i$ in Eq.~\eqref{eq:loss_id} is an one-hot indicator of 
the ground-truth identity of tracklet $\bm{T}_i$
and the $\text{FC}(\cdot)$ 
serves as a linear classifier 
which maps the tracklet's representation $\bm{x}_i$
into an identity prediction distribution $\tilde{\bm{y}}_i$
while $C$ is the total number of identities.
Moreover,
the triplet ranking loss $\mathcal{L}_{\text {trip}}$ 
explicitly draws the features of a positive tracklet pair sharing the same identity
closer in the learned latent space
while pushes the negative pairs apart:
\begin{equation}
\mathcal{L}_\text{trip}(\bm{T}_i) = \max(0, \Delta 
+ \mathcal{D}(\bm{x}_i, \bm{x}_i^+) 
- \mathcal{D}(\bm{x}_i, \bm{x}_i^-)),
\label{eq:loss_trip}
\end{equation}
where $\bm{x}_i^+$ and $\bm{x}_i^-$ are 
the representations of two randomly sampled tracklets 
with the same and different ground-truth labels as $\bm{x}_i$ in respective,
$\mathcal{D}(\cdot, \cdot)$ measures the distance of two features
and $\Delta$ is a predefined margin.
The overall optimisation objective of a batch of tracklets
is then formulated by 
combining the two losses as:
\begin{equation}
\mathcal{L} = \frac{1}{n}\sum_{i=1}^n 
(\mathcal{L}_\text{id}(\bm{T}_i) +
\mathcal{L}_\text{trip}(\bm{T}_i)),
\label{eq:loss_all}
\end{equation}
where $n$ is the size of a mini-batch.
Since the objective function Eq.~\eqref{eq:loss_all} is differentiable,
the \method{abbr} model can be trained end-to-end
by the conventional stochastic gradient descent algorithm
in the batch-wise manner.

\vspace{-0.5em}
\section{Experiments}
\vspace{-0.5em}

\begin{figure}
	\centering
	\includegraphics[width=\textwidth]{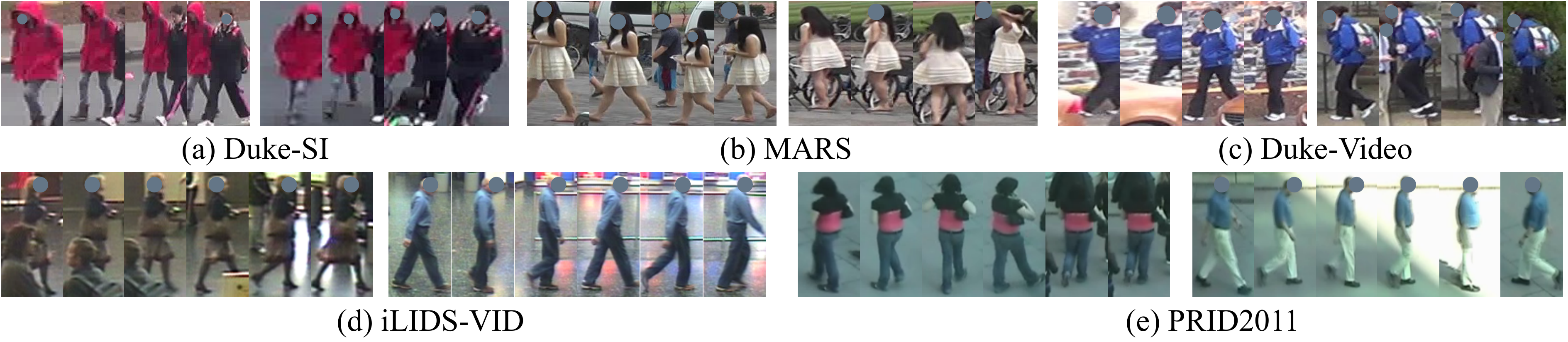}
	\vspace{-2em}
	\caption{Example pairwise tracklets with the same ground-truth identity labels. 
	Various noises are caused by illumination, viewpoints, resolution, occlusion, background clutter, etc.}
	\vspace{-1.5em}
	\label{fig:dataset_example}
\end{figure}

\paragraph{Datasets.} 
The proposed \method{full} (\method{abbr}) 
is evaluated on four video-based Re-ID datasets: 
MARS~\cite{zheng2016mars}, 
Duke-Video~\cite{ristani2016performance, wu2018exploit}, 
Duke-SI~\cite{li2019unsupervised}, 
iLIDS-VID~\cite{wang2014person},
and PRID2011~\cite{hirzer2011person}.
Example tracklets are shown in Fig.~\ref{fig:dataset_example}.
The MARS has 20,478 tracklets of 1,261 persons captured from a camera network with 6 near-synchronised cameras. 
Duke-Video is a newly released large-scale benchmark of 1,812 person identities with 4,832 tracklets. 
Duke-SI is a fully auto-generated version of Duke-Video
without manual frames selection,
thus, more practical and challenging.
The iLIDS-VID dataset 
is relatively small scale
including 600 video tracklets of 300 persons captured by two disjoint cameras in an airport arrival hall. 
The PRID2011 is another small scale dataset
containing 1,134 tracklets from 934 identities captured by two cameras.

\paragraph{Evaluation Metrics.} 
To evaluate the effectiveness of the proposed \method{abbr} model,
we adopted two commonly used performance metrics in person re-id including
Cumulative Matching Characteristics (\textbf{CMC}) 
and Mean Average Precision (\textbf{mAP})~\cite{zheng2015scalable}.

\paragraph{Implementation Details.} 
For fair comparisons,
we took a ResNet50~\cite{he2016deep} as the backbone network for global visual feature extraction~\cite{gu2020appearance}.
Given that the video tracklets are composed of arbitrary number of frames,
we split each tracklet into several clips with a fixed length of 10. 
We randomly sampled $4$ identity instances each with 8 clips to construct a mini-batch
in model training.
All the frames were resized to $256 \times 128$ and 
augmented by random horizontal flip. 
We used Adam~\cite{kingma2014adam} with weight decay of $5e-4$ for model optimisation. 
The margin $\Delta$ in Eq.~\eqref{eq:loss_trip} is set to 0.3, 
and the dimension $D$ of representations is set to $2048$ following~\cite{gu2020appearance, luo2019bag}. 
The kernel size $S$ for the 1D convolution in Eq.~\eqref{eq:laq_conv} is set to 10. 
The model was trained on two P100 GPUs for 240 epochs, 
and the learning rate is initialised to $3e-4$ 
which linearly decayed with a factor of $0.1$ per $60$ training epochs. 
During the testing stage, 
the tracklet-level representation was obtained by 
averaging pooling the learned representations of their clips. 
Cosine distance was then used to measure the distances between a query and every probed tracklet in gallery
for Re-ID.

\subsection{Comparisons to the State-of-the-Art}

\label{sec:SOTA}

\begin{table}[htbp]
  \centering
  \resizebox{\textwidth}{!}{%
    \begin{tabular}{l|cccc|cccc|cccc|ccc|crr}
\hline   \multirow{2}{*}{Methods} & \multicolumn{4}{c|}{Duke-Video} & \multicolumn{4}{c|}{Duke-SI}  & \multicolumn{4}{c|}{MARS}     & \multicolumn{3}{c|}{iLIDS-VID} & \multicolumn{3}{c}{PRID2011} \\
\cline{2-19}          & mAP   & R1    & R5    & R20   & mAP   & R1    & R5    & R20   & mAP   & R1    & R5    & R20   & R1    & R5    & R20   & R1    & R5 & R20 \\
    \hline
    TAUDL~\cite{li2018unsupervised} & -     & -     & -     & -     & 20.8  & 26.1  & 42.0  & 57.2  & 29.1  & 43.8  & 59.9  & 72.8  & 26.7  & 51.3  & 82.0  & 49.4  & 78.7  & 98.9 \\
    EUG~\cite{wu2018exploit} & 78.3  & 83.6  & 94.6  & 97.6  & -     & -     & -     & -     & 67.4  & 80.8  & 92.1  & 96.1  & -     & -     & -     & -     & - &  \\
    Snippet~\cite{chen2018video} & -     & -     & -     & -     & -     & -     & -     & -     & 76.1  & 86.3  & 94.7  & 98.2  & 85.4  & 96.7  & 99.5  & 93.0  & 99.3  & \best{100.0} \\
    VRSTC~\cite{hou2019vrstc} & 93.5  & 95.0  & 99.1  & 99.4  & -     & -     & -     & -     & 82.3  & 88.5  & 96.5  & -     & 83.4  & 95.5  & 99.5  & -     & - & - \\
    GLTP~\cite{li2019global} & 93.7  & 96.3  & \scnd{99.3} & \scnd{99.7} & -     & -     & -     & -     & 78.5  & 87.0  & 95.8  & 98.2  & 86.0  & \scnd{98.0} & -     & \scnd{95.5} & \best{100.0} & - \\
    UTAL~\cite{li2019unsupervised} & -     & -     & -     & -     & 36.6  & 43.8  & 62.8  & 76.5  & 35.2  & 49.9  & 66.4  & 77.8  & 35.1  & 59.0  & 83.8  & 54.7  & 83.1  & 96.2 \\
    STMP~\cite{liu2019spatial} & -     & -     & -     & -     & -     & -     & -     & -     & 72.7  & 84.4  & 93.2  & 96.3  & 84.3  & 96.8  & 99.5  & 92.7  & 98.8 & \scnd{99.8} \\
    STA~\cite{fu2019sta} & 94.9  & 96.2  & 99.3  & 99.6  & - & - & - & - & 80.8  & 86.3  & 95.7  & 98.1  & - & - & - & -     & - & - \\
    STAR~\cite{wu2019spatio} & 93.4  & 94.0  & 99.0  & \scnd{99.7} & -     & -     & -     & -     & 76.0  & 85.4  & 95.4  & 97.3  & 85.9  & 97.1  & \scnd{99.7} & 93.4  & 98.3  & \best{100.0} \\
    FGRA~\cite{chen2020frame} & -     & -     & -     & -     & -     & -     & -     & -     & 81.2  & 87.3  & 96.0  & 98.1  & 88.0  & 96.7  & 99.3  & \scnd{95.5} & \best{100.0} & \best{100.0} \\
    MG-RAFA~\cite{zhang2020multi} & -     & -     & -     & -     & -     & -     & -     & -     & \best{85.9} & 88.8  & \best{97.0} & \best{98.5} & \scnd{88.6} & \scnd{98.0} & \scnd{99.7} & \best{95.9} & \scnd{99.7}  & \best{100.0} \\
    AP3D~\cite{gu2020appearance} & \scnd{95.6} & \scnd{96.3} & \scnd{99.3} & \best{99.9} & \scnd{74.7} & \scnd{79.3} & \scnd{91.7} & \scnd{97.4} & \scnd{84.5} & \best{90.4} & \scnd{96.6} & \scnd{98.4} & 86.7  & 98.0  & \scnd{99.7} & 94.4  & 98.9  & \best{100.0} \\
    \hline
    \method{abbr} & \best{96.6} & \best{97.0} & \best{99.4} & \best{99.9} & \best{76.6} & \best{81.0} & \best{92.8} & \best{97.8} & 84.1  & \scnd{89.5} & 96.3  & 97.9  & \best{91.3} & \best{99.3} & \best{100.0} & \best{95.9} & 98.9  & \best{100.0} \\
    \hline
    \end{tabular}
    }
	\caption{Comparisons to the state-of-the-art video person Re-ID methods. 
	    Results of the prior methods are from the original papers {or reproduced by the official codes}.
	    The 1st/2nd best results are in \best{bold}/\scnd{underlined}.
	    `$\dagger$': unsupervised.}
	\label{tab:sota}
	\end{table}%

\noindent 
In Table~\ref{tab:sota},
we compared the proposed \method{abbr} model 
with a wide range of state-of-the-art video person Re-ID methods.
The \method{abbr} model yielded the best results across the board,
which suggests the efficacy of 
associatively exploring local part alignments and global appearance correlation 
in assembling a discriminative representation of a tracklet. 
Whilst maintaining its competitiveness
on the large-scale MARS and the well-curated Duke-Video datasets,
the \method{abbr} model achieved compelling improvements over the other methods on iLIDS-VID
and its performance advantage is more significant
on the automatically detected and segmented Duke-SI,
in which case \method{abbr} outperformed the others by 
1.9\%$\sim$55\%, 1.7\%$\sim$54.9\% and 1.1\%$\sim$50\% %
on mAP, rank-1 and rank-5, respectively. %

\vspace{-0.5em}
\subsection{Ablation Study}
\label{sec:ablation}

We conducted further studies 
to experimentally investigate 
the effectiveness of 
exploring the complementary local and global information
by solely considering one while ablating another, 
and also demonstrated the superiority 
of our associative assembling 
over the dual-branch strategy~\cite{chen2020frame}
which used both local and global information separately.  
We also provided comprehensive visualisation 
for intuitively understandings.

\begin{figure}
	\centering
	\includegraphics[width=\textwidth]{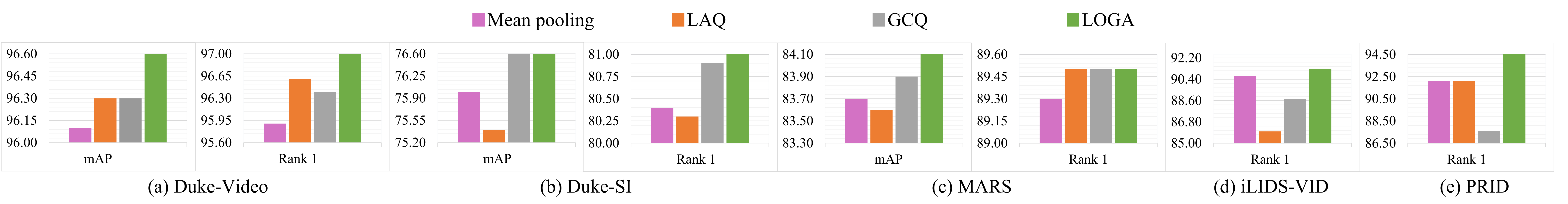}
	\vspace{-1.0em}
	\caption{Ablation studies on model components.}
	\label{fig:abl_comp}
	\vspace{-1.0em}
\end{figure}

\paragraph{Components analysis.}
We started with examining the role of 
local part alignments 
by introducing LAQ for frame assembling. 
Fig.~\ref{fig:abl_comp} (pink \vs orange) shows
that both metrics on most datasets are decreased. 
This is caused by the unrealistic assumption that
local regions of all the frames are well-aligned. 
Such an assumption is shown to be unreliable due to 
uncontrollable environment and fragile detection/segmentation.
We further examined the importance of global appearance 
by solely employing GCQ for frame assembling.
The unsatisfying performance 
as reported in Fig.~\ref{fig:abl_comp} (pink \vs gray)
suggests assessing the quality of frames 
in accordance with solely 
the unobstructed global appearance is unreliable
owing to the fine-grained details being ignored. 
In contrast, when both LAQ and GCQ are adopted, 
LOGA exhibits remarkable advantage 
over all other counterparts (green \vs others).
This demonstrates the indispensable of both LAQ and GCQ.

\begin{figure}[!htb]
   \begin{minipage}{0.50\textwidth}
     \centering
     \includegraphics[width=\textwidth]{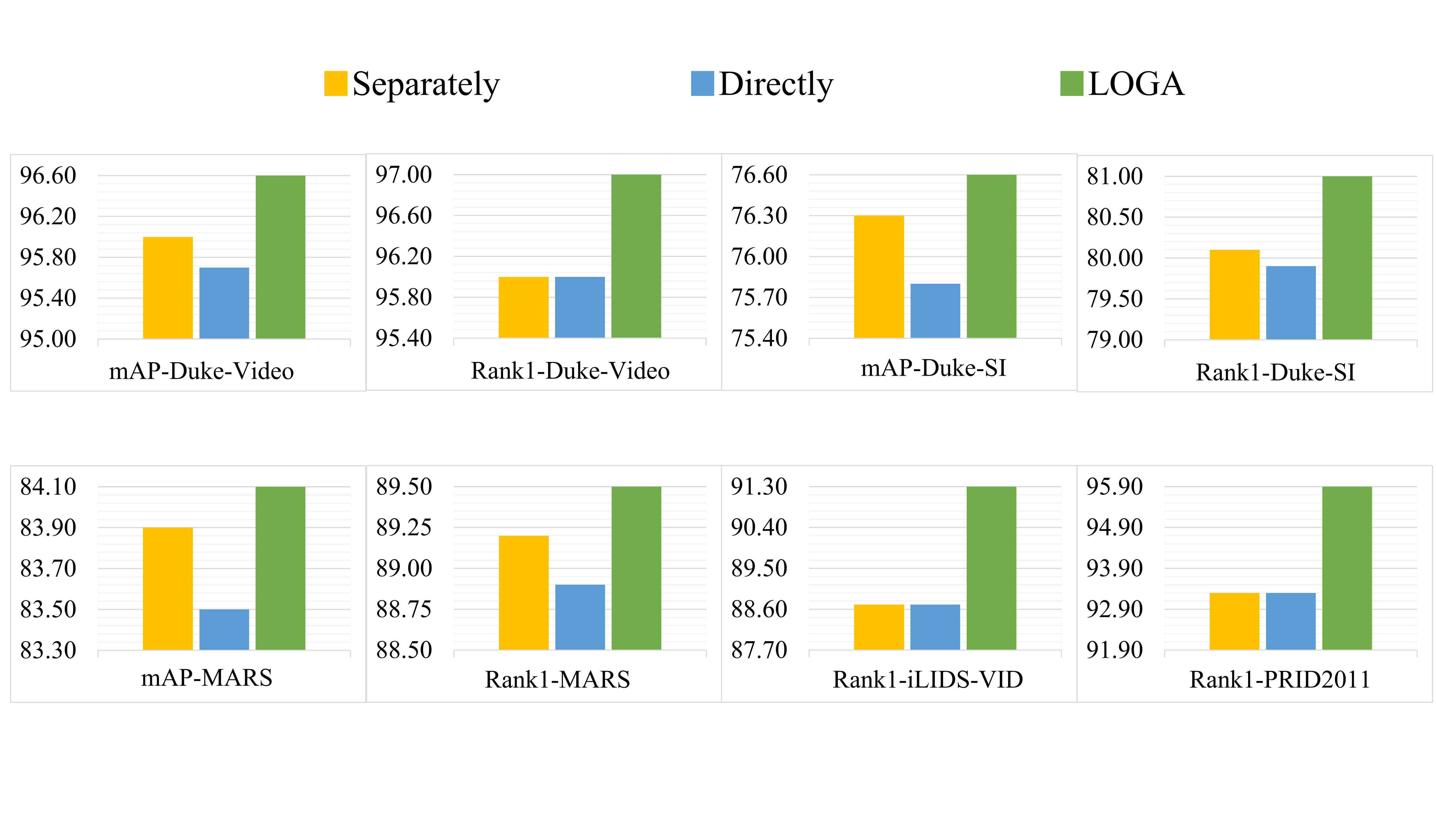}
          \vspace{-1em}\hfill
     \caption{Impacts of assembling strategies.}\label{fig:abl_asmb}
   \end{minipage}
   \begin{minipage}{0.50\textwidth}
     \centering
     \includegraphics[width=\textwidth]{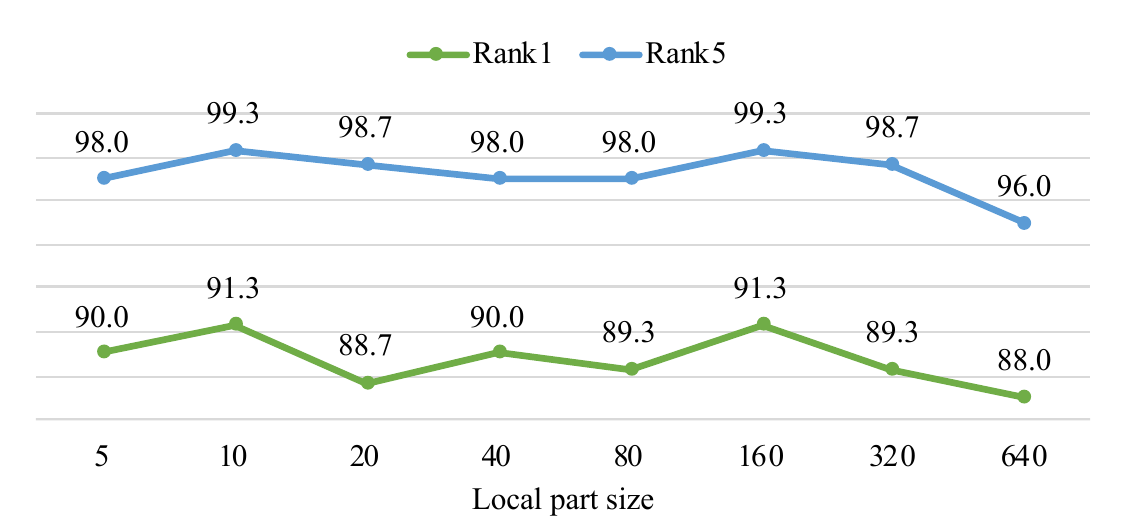}
          \vspace{-1em}
     \caption{Impacts of local part size in LAQ.}\label{fig:block_size}
   \end{minipage}
     \vspace{-1em}
\end{figure}

\paragraph{Effects of assembling strategy.}
We further studied the effects of different strategies
to join the local and global information in frames assembling:
(1) \textit{separately} assembling
by two individual branches learned in parallel 
according to the two kinds of information~\cite{chen2020frame}.
(2) \textit{directly} connecting 
local and global information 
by rescaling the per-frame visual features $\bm{E}$
according to their normalised local alignment scores (Eq.~\eqref{eq:laq_mlp})
then explore their global correlations by 
the conventional self-attention on the rescaled features.
(3) \textit{associatively} assembling 
by combining the local-assembled prototype
and global-assembled residual (Eq.~\eqref{eq:laq_gaq})
to exploit their synergy.
The comparison given in Fig.~\ref{fig:abl_asmb} (green \vs others) shows a noticeable advantage of \method{abbr} over the dual-branch or direct-connecting counterpart, which demonstrates the effectiveness of the proposed associative assembling strategy.

\paragraph{Effects of local part size.} 
We study the effects of local part size 
by varying the kernel size of the 1D convolution in Eq.~\eqref{eq:laq_conv}
and experimented on iLIDS-VID. 
The experimental results shown in Fig.~\ref{fig:block_size} indicates 
our model's robustness to this hyper-parameter 
within a wide range of values
thanks to the subsequent GCQ module
which help refine the local alignment scores
according to global correlations. 
Given that improving $S$ doesn't benefit the performance
but increase the model's complexity,
we set $S=10$ in practice.

\begin{figure}[H]
	\centering
	\includegraphics[width=\textwidth]{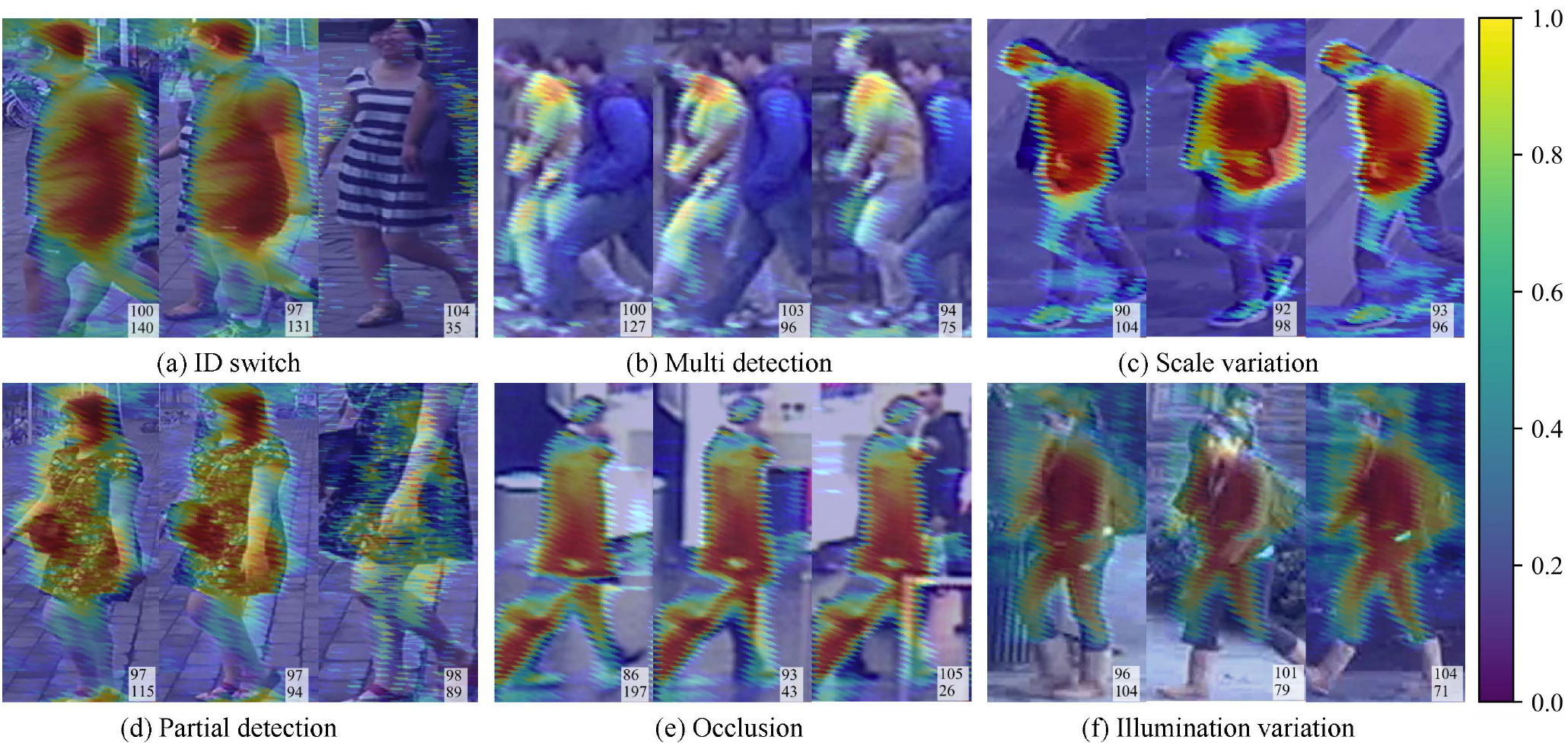}
	\caption{Visualisation of video clips suffering from various noise.  
    % The local-alignment scores and global-correlation scores
    % are shown at the upper and lower parts 
    % of each frame's right-bottom corn, respectively.}
	Their corresponding importance in assembling 
	are shown at the right-bottom corner of each frame
	with the local-alignment scores at top and the global-correlation scores
        beneath (amplified by 1,000 times). } %
	\label{fig:vis}

\end{figure}

\paragraph{Qualitative studies.} 
Fig.~\ref{fig:vis} shows several video clips
stacked with their activation maps 
generated according to their local parts quality.
Each frame's local-aligned score (upper, Eq.~\eqref{eq:laq_mlp})
and global-correlated score (lower, Eq.~\eqref{eq:gaq_scores})
are attached at their bottom-right corner.
% We visualised several video clips 
% with the corresponding quality assessment results
% including quality scores and activation map by local alignment assessment.
% for intuitively understandings of \method{abbr}.
% The white box located at the bottom-right corner of each frame in Fig.~\ref{fig:vis}
% indicates the local-aligned score (upper, Eq.~\eqref{eq:laq_mlp}) 
% and global-correlated score (lower, Eq.~\eqref{eq:gaq_scores}).
% As exhibited in Fig.~\ref{fig:vis}, 
As exhibited,
\method{abbr} is robust to various kinds of noise 
by providing a faithful importance score 
for assembling a discriminative representation. 
The activation maps accurately reveal the critical regions for Re-ID.
The global-correlated scores 
are obtained with the complementary appearance information 
so can reliably adjust the biased local-aligned scores.
For instance, as shown in Fig.~\ref{fig:vis}, 
LAQ enables network to focus on the target instead of the switched ID or the irreverent multi-detected ID as shown in the activation maps. For the low quality frames caused by partial-detection, scale-variation and occlusion, etc. LAQ can faithfully assess the local quality.
The suitable importance score revealed by the association of LAQ and GAQ efficiently guide LOGA to learn the representation from the most discriminative region in the most discriminative frames.

\section{Conclusions}
In this work,
we present a novel \textit{\method{full}} (\method{abbr}) method
for video person Re-ID through
selectively assembling video frames of diverse qualities 
to derive a more reliable and discriminative representation
of a video tracklet.
This is accomplished by assessing the frame's quality
according to both their \textit{local part alignments}
and \textit{global appearance correlation}
so to
refrain from integrating undesired visual information
into tracklet's representation causing 
identity mismatch.
Different from existing approaches
which explore either local or global information separately,
our \method{abbr} method constructs a 
local-assembled global appearance prototype of a tracklet
so to alleviate biased quality assessment 
caused by either identity-irrelevant misalignment
or spatial-insensitive appearance miscorrelation.
Extensive experiments on five benchmark datasets show the performance advantages
of \method{abbr} over
a wide range of the state-of-the-art video Re-ID methods.
Detailed ablation studies are also conducted
to provide in-depth discussions about
the rationale and essence of different components in our model design.

%\section*{Acknowledgements}
%This work was supported by Vision Semantics Limited, the Alan Turing Institute Turing Fellowship, and the China Scholarship Council.

\clearpage
\bibliography{egbib}
\end{document}